\pdfoutput=1
\documentclass[11pt]{article}

\usepackage[]{ACL2023}

\usepackage{times}
\usepackage{latexsym}
\usepackage{graphicx}
\usepackage{dialogue}
\usepackage[T1]{fontenc}
\usepackage[utf8]{inputenc}

\usepackage{microtype}

\usepackage{inconsolata}

\title{Response-conditioned Turn-taking Prediction}

\author{Bing'er Jiang \and Erik Ekstedt \and Gabriel Skantze \\
       Division of Speech, Music and Hearing,
       KTH Royal Institute of Technology\\
       \texttt{\{binger, erikekst, skantze\}@kth.se}}
\begin{document}
\maketitle
\begin{abstract}
Previous approaches to turn-taking and response generation in conversational systems have treated it as a two-stage process: First, the end of a turn is detected (based on conversation history), then the system generates an appropriate response. Humans, however, do not take the turn just because it is likely, but also consider whether what they want to say fits the position. In this paper, we present a model (an extension of TurnGPT) that conditions the end-of-turn prediction on both conversation history and what the next speaker wants to say. 
We found that our model consistently outperforms the baseline model in a variety of metrics. The improvement is most prominent in two scenarios where turn predictions can be ambiguous solely from the conversation history: 1) when the current utterance contains a statement followed by a question; 2) when the end of the current utterance semantically matches the response. 
Treating the turn-prediction and response-ranking as a one-stage process, our findings suggest that our model can be used as an incremental response ranker, which can be applied in various settings.

%Potential applications include interview settings where 
%Our findings further indicates that such mechanisms can be incorporated in interview settings where the response (interview questions) are predefined.

\end{abstract}

\section{Introduction}

A fundamental component of spoken dialog system (SDS) is turn-taking, i.e., the decision of when to take turns at appropriate places, without causing long response delays or interrupting the user. In other words, the system must be able to correctly identify when the user is yielding the turn, and it is appropriate to make a response, and when the user is simply making a mid-utterance pause \citep{Skantze2021}. Traditionally, this has been done using a simple silence threshold. However, silence is not a very good indicator of turn-shifts and more modern approaches instead use various cues known to be important in human-human turn-taking, such as lexico-syntactic cues, prosody, or gaze \citep{gravano101679,Ishii2016,Lala2019,ekstedt_how_2022}. 

\citet{turngpt} proposed TurnGPT, a transformer-based language model that incrementally processes words in the user's utterance and predicts the probability of a turn-shift after each word. This is similar to the notion of syntactic or pragmatic completion points that have been identified in conversation analysis \citep{ford1996}. In their analysis of TurnGPT, \citet{turngpt} found that the 20\% of the model's attention is directed towards utterances earlier than the current one, indicating that it is sensitive to pragmatic aspects of dialogue. 

While such models are indeed a step forward, there is a still an important component missing that we will address in this paper. When humans make a decision to take the turn, it is not just based on whether there are enough turn-yielding cues in the interlocutor's utterance. \citet{sacks1974} use the notion of transition-relevant places, or TRP, for places where a transition could potentially take place (but does not have to). Thus, many places for turn-shifts are highly optional. To partly address this problem, \citet{ishii_trimodal_2022} annotated the willingness of the next speaker to take the turn, and built a model that could predict this willingness based on multimodal cues. 

Whether a turn-shift takes place or not also depends on the intention of the next speaker, and what they want to say. For dialogue systems, this means that the system should not automatically take the turn once the transition-probability passes a certain threshold, and only then decide what it should respond. Instead, the system should take the potential response into account when deciding whether it is appropriate to take the turn or not. 
%In doing so, it might also take the utility of the response into account. If it has a response it ``wants'' to give (the utility is high enough), and which also fits into the current context in terms of turn-taking, it might respond, but otherwise it might be better to wait for further input. 

We call this \textbf{response-conditioned turn-taking prediction}, which is illustrated in Figure 1. In this paper, we investigate to what extent and under what scenarios such response-conditioning would help to predict turn-shifts. We present a model called \textbf{RC-TurnGPT}, which is an extension of TurnGPT. 

\begin{figure}[h]
  \centering
  \includegraphics[width=\linewidth]{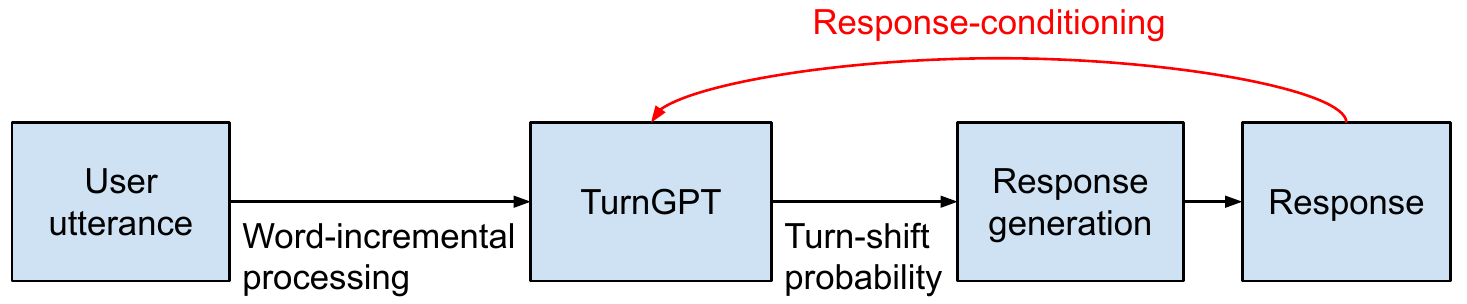}
  \caption{Response-conditioned turn-taking prediction.}
\end{figure}

Note that the current study does \textit{not} intend to address \textit{how} and \textit{when} the next speaker comes up with what they would like to say. This depends of course on the exact implementation of the dialogue system, which could for example be response-ranking \citep{gao-etal-2020-dialogue} or an intent-based planning approach \citep{meta_diplomacy_2022}. 
%Here, we take a first step in this direction. 
Regardless of this, the model proposed here could be used to incrementally rank or score potential responses to see whether they fit well from a turn-taking perspective. 
%, and analyse to what extent conditioning the prediction helps turn-taking. %that what extent ``knowing what to say'' helps turn-taking prediction. 

\section{Methods}

% unidirectional or causal ? unidirectional or causal-decoder based?
%*causal transformer decoder models \cite{transformer}
%*causal decoder-based transformer models \cite{transformer}

TurnGPT is a unidirectional transformer-based language model (LM) optimized through cross-entropy to predict the next token in a sequence. It is a pre-trained GPT-2 (base) model \citep{GPT}, finetuned on \textit{unpunctuated} dialog corpora, with a special turn-shift token (TS) that delimits consecutive turns. 
%The probability associated with the TS token is considered to be the probability of a TRP following the currently focused step. 
RC-TurnGPT is an extension of this model, by also conditioning the prediction on the response. 

% this turned in to almost an introduction paragraph...
%Because TurnGPT is unidirectional it can only infer a TRP given the preceding context and does not have any way of making its estimations on anything an agent plausibly would or could respond. This behavior is sub-optimal for turn-taking in general because the act of taking a turn is intrinsically linked with what to say, i.e. the probability of a TRP should be conditioned on the following response. Therefore, we extend the training objective of TurnGPT to introduce RC-TurnGPT, a unidirectional textual turn-taking model conditioned on both context and response to infer the probabilities of TRPs.

While the RC-TurnGPT model is architecturally equivalent to TurnGPT, it differs in the training objective through a simple data transformation. This transformation permutes the ordering of turns in a similar approach as the FIM pre-training objective of \citet{fim2022}. We consider turn-based dialog sequences to consist of three parts: the context/history (\textbf{H}), the current utterance (\textbf{CU}) and the next response (\textbf{R}). The task is to correctly predict the location of the turn-shift token in the current utterance, $CU_i$, given the history, $H_i$, and the next response, $R_i$, over all samples $i$ in the dataset, $D$. The samples $i \in D_I$ are extracted by applying a turn-based sliding window approach with a step size of 1 and a window size of 3 turns. 
% We should add that C could be multiple turns (of a max length?) and that R can be a turn?a sentence? max-tokens words? % I don't know the specifics now...

However, instead of the uniform left-to-right next token prediction task of regular LMs, the RC-TurnGPT model train on ordered sequences of \{R, H, CU\}, masking the loss over R and H to solely learn over the CU turns. This enables the model to use information of both H and R while keeping the original left-to-right next token prediction setup.

% I don't know if this is a suuuuper useful/important detail but here for completeness...
% Can be shortened but just writing it out...
Finally, the TurnGPT model utilized three special tokens in addition to the original GPT-2 vocabulary, the aforementioned TS token and two speaker tokens. The speaker tokens are similar to positional embeddings and are added to the word embeddings to encode the speaker identity over each word. Because of the permuted ordering of the RC-TurnGPT setup we also include a fourth special response-token that are added to the words of the response to distinguish them from the actual context. Both the base model and the datasets were implemented using Huggingface \cite{hf-trans, hf-dset}.

\subsection{Data}
We train RC-TurnGPT and the baseline TurnGPT on two types of data sets based on \citet{turngpt}: \textbf{Assistant} and \textbf{Written Social}. 
The former constitutes of three task-oriented dialog corpora: Taskmaster~\cite{taskmaster}, MetaLWOZ~\cite{metalwoz}, and MultiWoz~\cite{multiwoz}. The latter includes two corpora constructed by human-human written dialogs: CuriosityDialogs \cite{rodriguez2020curiosity} and DailyDialog~\cite{dailydialog}. 
All datasets are written dialogs with clearly defined turns. The resulting full dataset contains 106,830 dialogs for training, 9,362 for validation, and 7,897 for test, with an average number of turns being 13.69.
\subsection{Evaluation}
 % Erik: feel free to change text or the evaluation metric names

To evaluate the models, we propose five turn-level based metrics that measures the turn-shift performance in various ways. The models are considered to make a turn-shift prediction when the probability exceeds a certain threshold optimized for performance over the validation split, for each model independently.

First, we define turn-level accuracy (\textbf{TL-Acc}) to be the percentage of turns where the turn-shift probability exceeds the threshold at, and \textit{only} at, the ground-truth end of turn. %The threshold is determined by looping from 0 to 1 with an interval of 0.01, and the value that gives the highest TL-Acc is selected.
Second, the no response rate (\textbf{NRR}) is the percentage of turns where the threshold is never exceeded and the model fails to make a response. The third metric is defined to measure the barge-in rate (\textbf{BR}), the percentage of turns where the models would make a turn-shift prediction before the actual turn-shift. 

%In addition to the turn-shift based metrics, w
We also investigate instances where the two models make different turn-taking decisions to see how well the response would fit, using perplexity as a measure.  
%we imagine that the models would in fact have taken the turn, at their first turn-shift prediction, and responded with the subsequent turn. 
We use the TurnGPT model to calculate the average perplexity over the response (\textbf{R-PPL}). 
%This response perplexity (\textbf{R-PPL}) 
%encodes how likely the response is, at the predicted turn-shift, given the context. 
%Because the perplexity is the same over turns where the models prediction are equivalent we only consider turns where they differ. 

Lastly, we define the ordinal spike rate (\textbf{OSR}) to be the percentage of turns where the probability is the greatest at the end of the turn. This metric does not consider a threshold but simply measures how many times the highest probability is located at the correct turn-shift location.

\section{Results}

\subsection{Aggregate results}
Table \ref{tab:aggregate} shows that RC-TurnGPT performs better in all evaluations metrics, although the improvement is not large overall. While 55.77\% turn-level accuracy may not seem very high, it should be noted that even predictions different from ground-truth turn-shift can also be valid in everyday conversations, especially in long utterances where several completion points are likely. 
While the threshold-based binary metric is low, the probability-based OSR is much higher, indicating that the model is indeed able to detect end of turn reflected by assigning the highest probability.
Furthermore, the perplexity of the response also decreases, showing that when one or both of the two models make a mistake, the response fits better with the context for the turn-shifts RC-TurnGPT takes.
%makes the conversation more natural.
%threshold-based metrics forces a binary decision, which makes it even harder 
%We believe that turn-level accuracy alone cannot reflect model's 
%this task in nature may allow for , given that there might be several transition-relevant places in one an the same utterace. 

\begin{table}[ht]
\centering
% Horizontal
% \begin{tabular}{ lrrrrr } 
% \hline
% Model & TL-acc &NRR&BR&R-ppl&OSR\\  
% \hline
% TurnGPT    & 53.93\% &- &- & 20.90\%-\\
% RC-TurnGPT & 56.93\% &- &- & 18.36\%-\\
% \hline
% \end{tabular}

% Vertical
\begin{tabular}{ lrr } 
\hline
Metric & Turn-GPT & RC-TurnGPT \\
\hline
TL-Acc $\uparrow$ & 53.93\% & \textbf{55.77\%} \\
NRR $\downarrow$   & 20.90\%       & \textbf{19.23\%} \\
BR $\downarrow$    & 25.17\% & \textbf{24.75}\% \\
R-PPL $\downarrow$ & 1.923       & \textbf{1.918} \\
OSR $\uparrow$   & 88.57\%      & \textbf{89.17}\% \\

\hline
\end{tabular}
\caption{The turn-level accuracy (TL-Acc), no response rate (NRR), barge-in rate (BR), response perplexity (R-PPL) and the ordinal spike rate (OSR) performance for TurnGPT and RC-TurnGPT. Best performance are bold. }
\label{tab:aggregate}
\vspace{1em}
\end{table}
\vspace{-2em}

\subsection{Model analysis}
In order to better understand \textit{when} conditioning on the response helps turn-shift prediction and when it does not, we proceed to analyse cases where only RC-TurnGPT makes the correct prediction, and where both models are successful.
%The aggregate results do not show a striking difference between the models, however, this can partly be explained by the the fact that many responses are quite generic and simple, which 
% Please check the part below... It is need of improvement
%On the other hand, sometimes the current utterances do not change in any meaningful way throughout the turn and a certain response can make sense even before the actual turn-shift. 

We extract all turns in the test set where TurnGPT makes a pre-mature turn-shift prediction but RC-TurnGPT correctly predicts the end of the turn. We sort the turns by the difference in probability assigned by the two models at the TurnGPT-predicted turn-shift. We then investigate the difference between the top and bottom 1000 cases. By comparing these two subsets, we can better understand when conditioning on the response makes the biggest difference. We identified two scenarios which we hypothesized would be important:
%We investigate two possible scenarios:
%In the following section, we analyse cases where conditioning on the response makes a difference, as well as cases where it does not.
% A closer investigation of the model comparison shows that RC-TurnGPT outperforms the baseline TurnGPT mostly in two scenarios: 
%1) multiple utterance; 2) semantic matching. We explain the two types using dialogues from the test set.
1) statement to question; 2) semantic matching. 
%We explain the two types using dialogues from the test set.

% Multiple "dialog acts" or "statement  to question"

%\paragraph{Multiple utterance} scenario refers to cases where the current utterance consists of more than one utterance, and the response is most relevant to the last utterance. Consider the following dialogue:

\begin{figure}[ht]
\begin{center}

\includegraphics[scale=0.65]{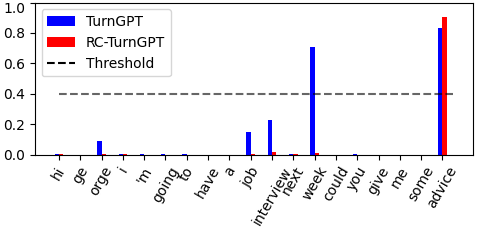}
\caption{Different turn-taking predictions: TurnGPT predicts the turn-shift at the end of a statement; RC-TurnGPT predicts the end of a question. Response: \texttt{sure first of all it's very important for you not to be late}} \label{fig:george}
\end{center}
\end{figure}

% Could be performed with a question classifier as well...
\paragraph{Statement to question} refers to cases where the current utterance consists of at least one statement and ends with a question. 
As there are more than one natural completion point, TurnGPT will be greedy while RC-TurnGPT will take the response into consideration and choose a later completion point as turn shift.
%Such turns put very different expectations on the nature of the response depending on where the turn-shift is predicted. 
%Provided a scenario where the RC-TurnGPT model is constrained to respond with a set of answers to questions it has the ability to wait for the appropriate moment to answer the user. 
Consider the following dialogue in Figure \ref{fig:george} (Current Utterance plotted, Response in caption):

%\begin{dialogue}
%\speak{A}
%    Hi George, I'm going to have a job interview next week. Could you give me some advice?
%\speak{B}
%    Sure. First of all, it's very important for you not to be late. Job interviewers [...]
%\end{dialogue}

%Spk1: hi george i'm going to have a job interview next week/ could you give me some advice

%Spk 2: sure first of all it's very important for you not to be late job interviewers [...]

%Figure \ref{fig:george} shows that without conditioning on the response, TurnGPT spikes at the end of the first utterance, which indeed could be a relevant place to take the turn. However, as the response clearly corresponds to some request, RC-TurnGPT has much lower probability at the same position, and only spikes at the end of the last utterance.

Figure \ref{fig:george} shows that without conditioning on the response, TurnGPT spikes at an early completion point interrupting the current speaker. However, as the response clearly corresponds to an answer to a request, RC-TurnGPT waits until the speaker finishes their request.

In order to quantify this effect, 
%we calculate how often TurnGPT makes a mistake by missing a question, 
we use punctuations to calculate how often TurnGPT makes a mistake by missing a question.
We use the top/bottom subsets and ask GPT3\footnote{Model version: ``text-curie-001''} \cite{gpt3} to insert punctuation over the ground truth turns (\texttt{advice} in this example) and the incomplete TurnGPT predicted turns (\texttt{week} in this example).
We then calculate the ratio of cases where the former ends with a question mark while the latter does not. 
%We sort the turns by the TRP-probability difference assigned at the TurnGPT predicted turn-shift and compare the top and bottom 1000 cases. 
% this 'conclusion' is not that strong/great perhaps...
The top cases contain 36.3\% statements to questions and the bottom 11.7\%. The higher ratio in the top cases indicates that the RC-TurnGPT model recognizes this pattern and uses the response conditioning to wait for the appropriate moment to take the turn.

\begin{figure}[ht]
\begin{center}
\centering
\includegraphics[scale=0.65]{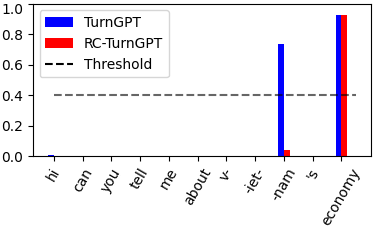}
\caption{Different turn-taking predictions: RC-TurnGPT's prediction allows closer semantic matching between current utterance and response. Response: \texttt{sure vietnam achieved an 8\% gdp growth between 1990 and 1997}} \label{fig:economy}
\end{center}
\end{figure}

\paragraph{Semantic matching} refers to cases where the response semantically corresponds to the specification made in the later parts of the current utterance. Consider the dialogue in Figure~\ref{fig:economy}:
% Spk1: can you tell me a little bit about vietnam's economy
% Spk2: sure vietnam achieved an 8\% gpd growth between 1990 and 1997

%\begin{dialogue}
%\speak{A}
%    Can you tell me a little bit about Vietnam's economy?
%\speak{B}
%    Sure! Vietnam achieved an 8\% GDP growth between 1990 and 1997.
%\end{dialogue}

\vspace{-0.7em}

As the response clearly addresses the topic of economy, Figure~\ref{fig:economy} shows that RC-TurnGPT would spike only after \texttt{economy} is specified, whereas TurnGPT has two spikes at both places and would predict the turn shift after \texttt{v-iet-nam}. It is important to note that while the response has no lexical overlap, the model still manages to find the semantic correlation.

In order to investigate whether RC-TurnGPT consitently recognizes such pattern, we use Sentence-Bert \cite{sbert} to measure the Semantic Textual Similarity between the Response and the last part of the actual turns missed by TurnGPT (here, \texttt{'s economy}). The average cosine distance for the top and bottom subsets are 0.293 and 0.209 respectively. 
% I don't know what we should say here below... PLEASE CHECK THIS
This indicates that where RC-TurnGPT outperforms TurnGPT, it does consider the semantic content of the response and delays predicting a turn-shift until the relevant semantic information has been stated. 
% This indicates that knowing the response is useful when semantically relevant information is uttered in the latter part of a turn.

\begin{figure}[ht]
\begin{center}

\includegraphics[scale=0.7]{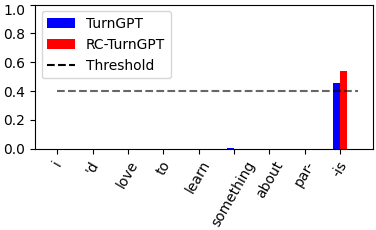}
\caption{Similar turn-taking predictions for a simple utterance. Response: \texttt{it is the capital of france}}\label{fig:other}
\end{center}
\end{figure}

\paragraph{Non-ambiguous turn-completions.} In addition, there are also a large number of cases where the current utterance has a fairly simple structure and hence it is not ambiguous where to take the turn.% regardless of knowing the response or not. 
In those cases, conditioning on the next response obviously makes a very small difference. As illustrated in Figure \ref{fig:other}, given that there is only one completion point, both models predict the turn shift correctly. This also explains why there are no drastic improvements for RC-TurnGPT when looking at aggregate results on the whole test set, as most of the task-oriented dialogues contain such simple utterances, which TurnGPT can perform well on. 
% Spk1: do you have any favorite places/ you have in mind
% Spk2: anywhere in colorado wyoming montana
%\begin{dialogue}
%\speak{A}
%    Do you have any favorite places/ you have in mind
%\speak{B}
%    Anywhere in Colorado, Wyoming, Montana
%\end{dialogue}

%\vspace{-2em}
\section{Discussion and conclusion}
In this study, we examined how turn-taking prediction can be improved when conditioned on the response. We found that the response conditioning is particularly helpful under two circumstances, mainly by preventing greedy turn-taking at earlier completion point: 1) when the current utterance contains statements followed by questions; 2) when the end of the current utterance semantically matches the response. 
However, for simple utterances with fewer completion points, TurnGPT is already capable of predicting the correct turn shift, and there is no additional help from conditioning on the response. %It is likely that given more challenging datasets, RC-TurnGPT will further outperform TurnGPT.

We should again stress that this paper does not address the question of how and when the system comes up with a potential response. 
%While we did not address how the speaker comes up with the response and we acknowledge that it is a nontrivial issue, 
However, this analysis shows that it is indeed possible to find a more suitable transition-point, when conditioning on the response. As we have suggested, the decision \textit{what} to say and \textit{when} it say it should be considered as a joint decision rather than a two-step process. In this regard, the RC-TurnGPT model could be used as an \textit{incremental response ranker}, which does not only consider different responses at each step, but which can also decide \textit{not} to respond and wait for more input. 
%we would like to point out that our study nevertheless sheds light on the mechanism of turn-taking prediction and response ranking as an one-step process. It can be seen and applied as both a turn prediction model and an incremental response ranker. 
For instance, it can be applied in an interview setting where the model (interviewer) asks questions (ranking from a list of interview questions) and take the turn at appropriate places.
%Our results indeed show the semantic matching between the response and the current utterance, suggesting a promising implementation incorporating the interview questions in a model-as-interviewer scenario. 
%Our future work will further apply utility to interview questions to further reflect human's intention of asking various questions.
For future work, it would also be interesting to involve the \textit{utility} of the candidate responses (from the system's perspective). In the interview scenario, this could for example mean that the system can find moments where certain important questions can be asked, and which also fit well from a turn-taking perspective. 

\section*{Limitations}
As mentioned above, the current study is limited to the question of whether (and when) conditioning turn-taking prediction on the response improves the performance. It does not yet show how the model could be incorporated in a spoken dialogue system. 
%this study does \textit{not} address \textit{how} and \textit{when} the next speaker comes up with what they would like to say, as this is beyond the scope of this short paper, and merits future work.
Moreover, this study focuses only on written conversations without incorporating spoken dialogues. Thus, the interpretations can be limited to dialogues that are relatively `formal' without hesitations, repetitions, etc.
Note also that we only analyse lexical cues to turn-taking (just like with TurnGPT), and leave out other modalities for future work. 

\section*{Ethics Statement}
%This study combines turn-taking prediction and response ranking as a one-step process, which is more similar to the cognitive processes in humans when they conduct conversations. 
%Our model can potentially be used to help understand human's cognitive processes of turn taking, by comparing the model judgements with human judgements in the same experiment. Through comparing the differences in performances, we can gain insights in how humans are able to conduct conversations in a smooth manner. Such research would also promote interdisciplinary collaboration.
The current study does not involve any human subjects and we do not foresee any ethical consequences.

\bibliography{custom}
\bibliographystyle{acl_natbib}

\appendix

%\section{Example Appendix}
%\label{sec:appendix}

%This is a section in the appendix.

\end{document}